\documentclass[conference]{IEEEtran}
\usepackage{times}

\usepackage[numbers]{natbib}
\usepackage{multicol}
\usepackage[bookmarks=true]{hyperref}
\usepackage{xcolor}
\usepackage{graphicx}
\usepackage{booktabs}
\usepackage{algorithm}
\usepackage{algpseudocode}
\usepackage{float}
\usepackage[font=footnotesize,labelformat=simple]{subcaption}
\usepackage[font=footnotesize]{caption}

\usepackage{amsmath,amsfonts,bm, bbm}

\def\eqref#1{equation~\ref{#1}}

\def\1{\bm{1}}

\DeclareMathAlphabet{\mathsfit}{\encodingdefault}{\sfdefault}{m}{sl}
\SetMathAlphabet{\mathsfit}{bold}{\encodingdefault}{\sfdefault}{bx}{n}

\DeclareMathOperator*{\argmax}{argmax}
\DeclareMathOperator*{\argmin}{argmin}

\newcommand{\xxnote}[3]{}
\ifx\hidenotes\undefined
  \renewcommand{\xxnote}[3]{\color{#2}{#1: #3}}
  \definecolor{SeaGreen}{HTML}{3FBC9D}
  \definecolor{Orange}{HTML}{D95F02}
  \definecolor{RacingGreen}{HTML}{004225}
\fi

\begin{document}

\title{Long Range Navigator (LRN): Extending robot planning horizons beyond metric maps}

\IEEEoverridecommandlockouts
\author{
    Matt Schmittle\textsuperscript{1*}\thanks{\textsuperscript{*}\texttt{\{schmttle, rbaijal\}@cs.washington.edu}},
    Rohan Baijal\textsuperscript{1*},
    Nathan Hatch\textsuperscript{3},
    Rosario Scalise\textsuperscript{1},\\
    Mateo Guaman Castro\textsuperscript{1},
    Sidharth Talia\textsuperscript{1},
    Khimya Khetarpal\textsuperscript{2,4},
    Byron Boots\textsuperscript{1},
    Siddhartha Srinivasa\textsuperscript{1}\\[1ex]
    \textsuperscript{1}University of Washington \hspace{5mm}
    \textsuperscript{2}Google DeepMind \hspace{5mm}
    \textsuperscript{3}Overland AI \hspace{5mm}
    \textsuperscript{4}Mila\\
    \textsuperscript{*}Equal Contribution
}

\maketitle

\begin{abstract}
A robot navigating an outdoor environment with no prior knowledge of the space must rely on its local sensing to perceive its surroundings and plan. This can come in the form of a local metric map or local policy with some fixed horizon. Beyond that, there is a fog of unknown space marked with some fixed cost. A limited planning horizon can often result in myopic decisions leading the robot off course or worse, into very difficult terrain. Ideally, we would like the robot to have full knowledge that can be orders of magnitude larger than a local cost map. In practice, this is intractable due to sparse sensing information and often computationally expensive. In this work, we make a key observation that long-range navigation only necessitates identifying good frontier directions for planning instead of full map knowledge. To this end, we propose the Long Range Navigator (\texttt{LRN}), that learns an intermediate affordance representation mapping high-dimensional camera images to `affordable' frontiers for planning, and then optimizing for maximum alignment with the desired goal. The \texttt{LRN} notably is trained entirely on unlabeled ego-centric videos making it easy to scale and adapt to new platforms. Through extensive off-road experiments on Spot and a Big Vehicle, we find that augmenting existing navigation stacks with \texttt{LRN} reduces human interventions at test-time and leads to faster decision making indicating the relevance of \texttt{LRN}. {\hypersetup{hidelinks}%
\href{https://personalrobotics.github.io/lrn/}{\textcolor{blue}{personalrobotics.github.io/lrn}}}

\end{abstract}%

\IEEEpeerreviewmaketitle

\section{Introduction}
\label{sec:introduction}

Autonomous off-road mobile robots require long-range waypoint navigation, often in environments where prior information (e.g. satellite imagery) is inaccurate or unavailable. Our goal is efficient navigation in these large-scale scenarios where our domain is significantly (10X or more) larger than the robot's sensor footprint. The central research question is, 

\begin{center}
\textit{How can we enable robots to make less myopic decisions facilitating \emph{long-range navigation} with only \emph{incomplete} knowledge of the environment?}
\end{center}

We assume that the robot has GPS localization together with target waypoints. Notably, the robot is provided no other information about the environment and must find its own path to the goal. The robot must achieve the goal with minimal human intervention and as fast as possible. Mobile robots with no prior information traditionally create a metric cost map based on onboard sensory information (e.g. cameras, lidar, and odometry)~\cite{meng2023, patel2024}. In an ideal setting, the local range of the map would suffice for short horizon of target locations, but with long-range goals, creating large cost maps is intractable due to limited range of sensors together with compute and memory limitations.

In particular, depth information required to project features into the map is often sparse and noisy. This results in a limited horizon of the cost map and the area outside this horizon being a fog of unknown space.

\begin{figure}
    \centering
    \includegraphics[width=1.0\linewidth]{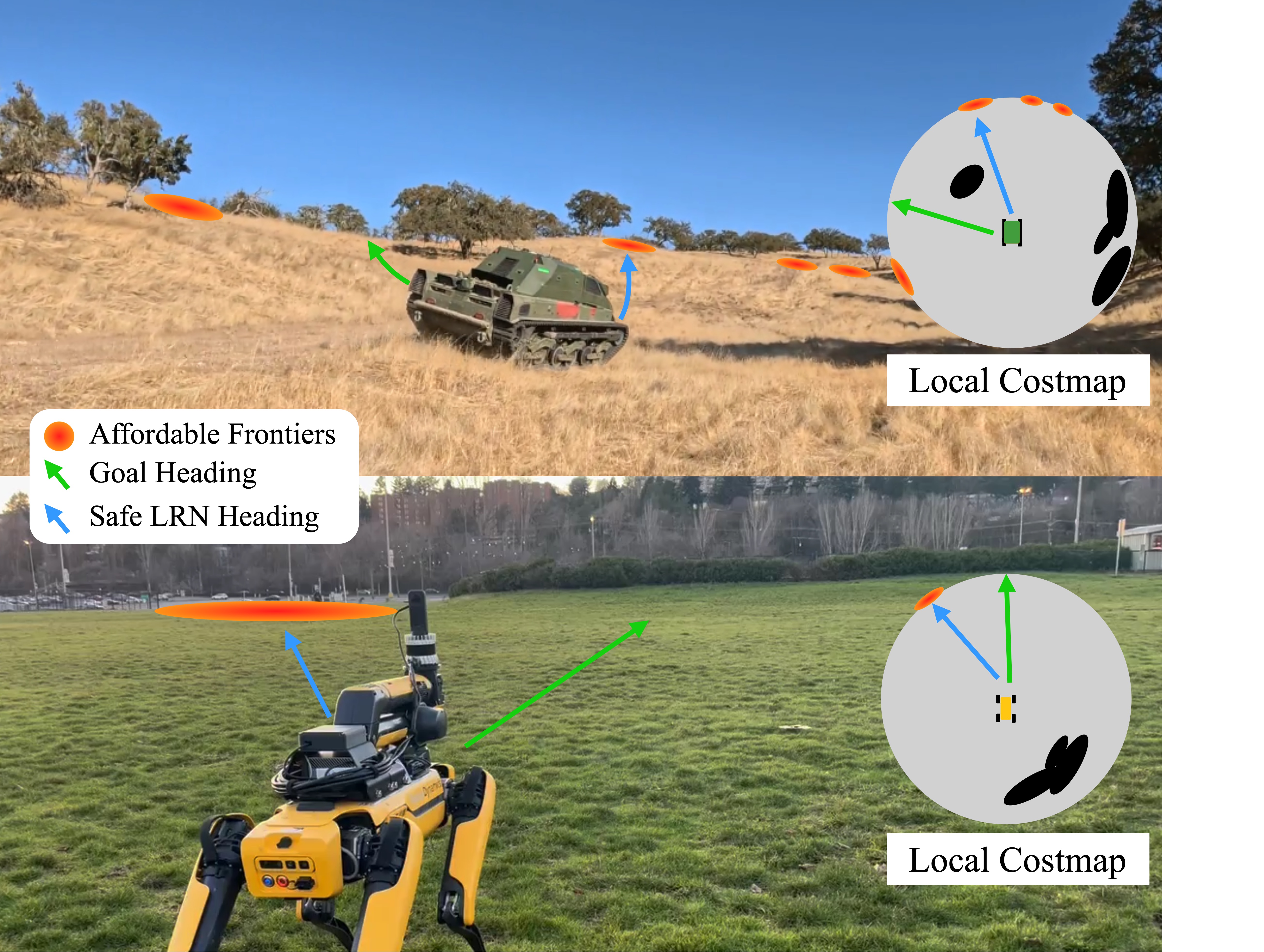}
    \caption{\textbf{\texttt{LRN} Overview.} Our approach \texttt{LRN} finds affordable frontiers as intermediate representation for the robot to head towards and selects one near the goal heading. On the right is the local perception (TOP 50m, BOTTOM 8m) where \texttt{LRN} changes the default navigation direction (green) to an affordable one (blue).}
    \label{fig:cover}
    \vspace{-20pt}
\end{figure}

\begin{figure*}[ht!]
    \centering
    \includegraphics[width=\linewidth]{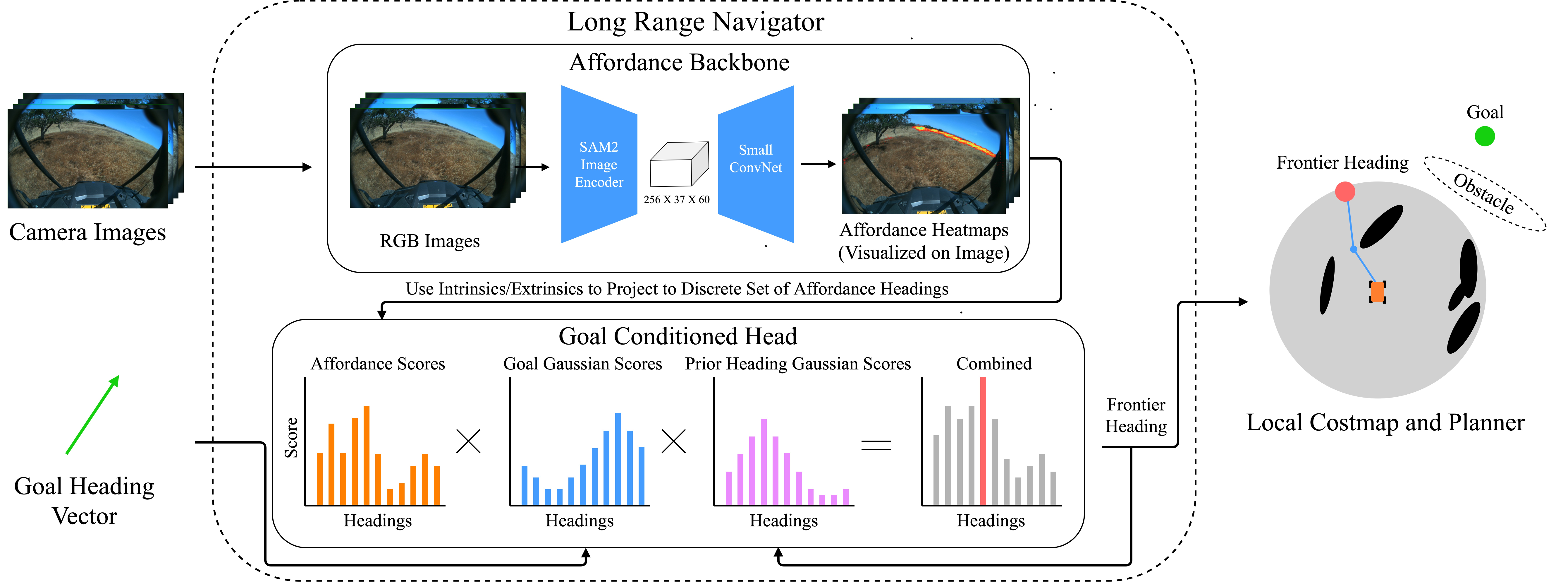}
    \caption{\textbf{Overview of our approach \texttt{LRN}}. \texttt{LRN} is fed with egocentric camera images and a goal heading vector. \texttt{LRN} is composed of the following components, namely, 1) \textbf{the Affordance Backbone}: computes \emph{affordable} frontiers in the image space as heatmaps agnostic of the goal. These affordance hotspots are then projected into a discrete set of affordable headings for the robot to follow, 2) 
    \textbf{the Goal Conditioned Head}, wherein the affordance scores are multiplied with a discrete gaussian score around the goal and a separate gaussian around the previous prediction (to maintain consistency). The maximum combined score heading (red) is selected. The local system can then use that frontier as a goal for local planning instead of the true goal. This process then repeats as new sensor information comes in.}
    \label{fig:bilevel}
    \vspace{-10pt}
\end{figure*}
A go-to approach to deal with unknown space is to heuristically assign it a fixed cost effectively planning straight to the goal once outside the map~\cite{patel2024}. The challenge is that in many scenarios, this leads to highly inefficient and potentially unsafe paths. A \textbf{key observation} is that field operators can determine the robot's long range strategy by simply analyzing the robot's image feed, without the requirement of a complete terrain map. For example, as seen in Fig.~\ref{fig:cover}, it is possible to see the opening in a wall of trees from images without mapping every tree. We refer to the boundaries between known/unknown regions as \emph{frontiers}. Frontiers which visually appear open i.e. possible to navigate to and continue beyond are referred to as \emph{affordable} frontiers. With this, we offer the following \textbf{key insight}: 
\begin{center}
    \textit{A robot can reason further out by learning to identify distant \emph{affordable} frontiers as intermediate goals.}
\end{center}

We propose improvements to current heuristic-based approaches by learning affordable frontiers in the image space. Specifically, we encode camera sensor images using the SAM2 foundation model~\cite{ravi2024sam2}, which is used to train a small decoder to predict long-range affordance heatmaps in a \emph{goal-agnostic} manner. We then project the heatmaps into heat scores for different headings the robot could follow and re-weight each heading based on the goal context. The top scoring direction is passed to the local system as a heading to follow to reach the goal. We refer to our method as \textbf{Long Range Navigator (\texttt{LRN})} depicted in Fig.~\ref{fig:bilevel}. \textbf{Key contributions} of our work are:
\begin{itemize}

    \item We introduce \texttt{LRN}, a long-range navigation approach with high-dimensional camera data to extend the planning horizon of current navigation systems. Our key insight is to leverage an intermediate affordance representation, which is sufficient for the agent to generate a long-horizon plan.
    \item To reduce the dependency on human-expert data annotation, we leverage video point tracking (CoTracker~\cite{karaev2023cotracker}) to automatically label data from ego-centric human walking videos. 
    \item We demonstrate the efficacy of \texttt{LRN} in real world outdoor navigation tasks, on a quadruped Spot robot and a Racer Heavy traveling distances of over a kilometer.
\end{itemize}
\section{Problem Setup}
\label{sec:formulation}

We study the long-range navigation problem setting where the robot perceives its environment from its local sensors (e.g. camera, lidar), and is tasked with navigating to a distant goal $g \in \mathcal{G}$ in a static, unknown environment. The robot is equipped with access to a local policy $\pi_\lambda: \mathcal{O} \rightarrow \mathcal{A}$, that maps the current observation $o$ to primitive actions $a \in \mathcal{A}$ and plans under a cost function $C : (s,a,s') \rightarrow \mathbb{R}$. Planning is limited to some horizon $H$ due to sensor or compute limitations. We assume the policy receives an observation $o$, plans, executes an action, and replans at some frequency in a model predictive control fashion. As it executes actions, the robot creates a path $\xi^\pi$. The \textbf{objective} is to minimize the expected cost of navigating to the goal in an unknown environment $\phi$. 
\[
J(\xi^\pi) := \argmin_{\xi^\pi} \mathbb{E}_{\phi\sim\mathcal{\Phi}}\Bigr[C(\xi^\pi)\Bigr]
\]

\section{\textbf{Long Range Navigator (\texttt{LRN})} }
\label{sec:approach}
\begin{figure}[h]
    \centering
    \includegraphics[width=0.6\linewidth]{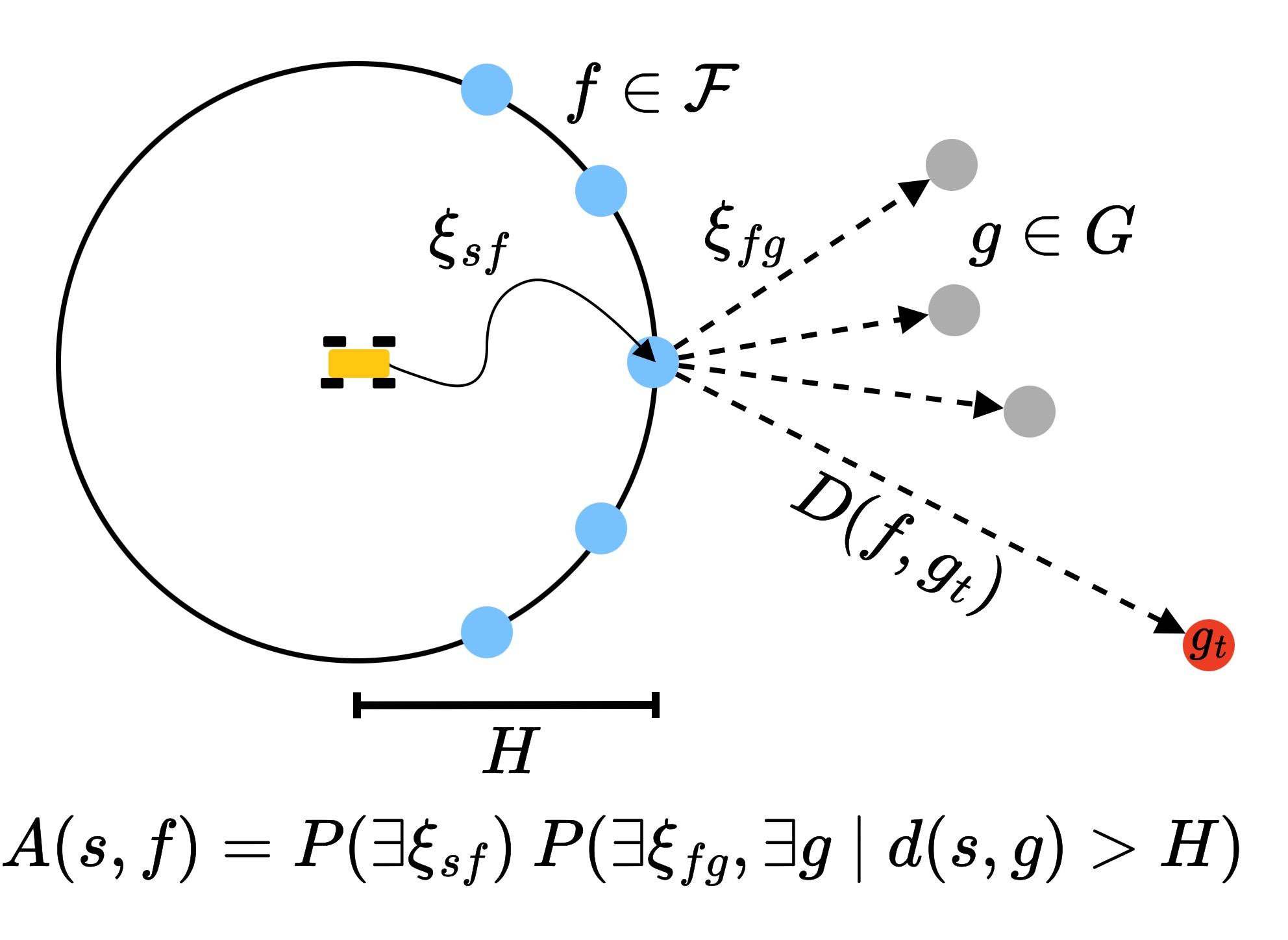}
    \caption{\textbf{\texttt{LRN}'s formulation of the long-range navigation problem.} \texttt{LRN} learns the value estimate $V(s, g, f)$ using affordability score $A(s,f)$ for each frontier and the cost to goal estimate $D(f, g_t)$.}
    \label{fig:problem}
\end{figure}

Core to our approach is the idea that while low-level controller policy $\pi_\lambda$ is limited to a local horizon, it does not mean that useful sensor data does not exist beyond that horizon for navigation. Consider a set of frontier states $f \in \mathcal{F}$ at the periphery of horizon $H$ which borders known/unknown space as shown in Fig.~\ref{fig:problem}. Under optimal substructure property~\cite{cormen2009}, if we know the optimal frontier node $f^*$ which lies on the optimal path from start to goal and $\pi_\lambda$ is optimal up to $H$ then $\pi_\lambda$ planning to $f^*$ is acting globally optimally. In practice, this is very hard because $\pi_\lambda$ will usually be sub-optimal and $f^*$ depends on unknown information not detected in any of the robot's sensors (e.g. backside of a hill). That being said, there may exist information within the robot's sensor range that can help estimate affordable frontiers which seem possible to navigate to and continue beyond.

To estimate affordable frontiers from state $s$, we first define the value of a frontier $f$ given a goal $s$ as $V(s,g,f)$. This can be decomposed into two parts, namely $A(s, f)$ and $D(f, g_t)$ as shown in Fig.~\ref{fig:problem}. For clarity a specific navigation goal at time $t$ is denoted $g_t$. Formally,
\begin{align}
V(s,g,f) &= A(s, f)\: D(f, g_t)\\
A(s, f) &= P(\exists \xi_{sf})\:P(\exists\xi_{fg}, \exists g \mid d(s, g) > H),
\end{align}

where $A(s,f)$ measures the \emph{affordability score}. This score is computed by multiplying the probability there exists a path from $s$ to $f$ and the probability there exists some path from $f$ to some distant goal $g$ beyond the local horizon $H$. $D(f, g_t)$ measures the  cost estimate of navigation from the frontier $f$ conditioned on the goal. Given, $V(s,g,f)$ and $\pi_\lambda$ we can define the policy we seek $\pi$.

\begin{align}
f_\pi = \argmax_{f \in F}V(s, g, f)\\
\pi(s, g) = \pi_\lambda(s, f_\pi)
\end{align}

\texttt{LRN} is a bi-level system, with two components: one to estimate $A(s) = \left[ A(s, f) \right]_{f \in F}
$ and the other to estimate $D(f, g_t)$. We implement the first component $A(s)$ to estimate \emph{all affordance scores} via a learned mapping from camera sensors to affordances (selective attention) in the image space. The second component scores frontier states given a goal context. The overall algorithm is depicted in Alg.~\ref{alg: lrn}. In practice, we discretize the space around the robot into angular bins which constitutes the space of frontiers $\mathcal{F}$.

Evaluating frontiers using image data can be challenging, due to potential projection errors or occlusions when trying to project frontiers into the image space. Besides, frontiers may not clearly associate with important features in the image. For example, if there exists a distant opening in the trees but the local costmap might not yet reach the treeline. The frontier may be in an open area far from the treeline, so evaluating its affordability will require relating it to information far from the frontier in image space. For these reasons, we propose instead to learn an intermediate representation of affordable image frontiers and project them to local frontiers.

Image frontiers are converted to frontiers by projecting them to a ray using camera intrinsics (we assume no reliable depth is available) and selecting the point at a distance $H$ along that ray. The projected image frontier is not where the image point truly is in 3D space. So to make this projection reasonable from a navigation standpoint the affordable image frontier must have a clear line of sight path from the projected point at distance $H$ to the true 3D point associated with the image frontier. This property is impossible to enforce perfectly without a prior map. Instead, we approximate it by learning a mapping from images to affordance heatmaps via automatic data labeling and human labels.

\begin{algorithm}
\caption{\texttt{LRN}: Long Range Navigator}
\begin{algorithmic}
\Require $k$ angular frontier bins, initial state $s_{start}$, goal $g$, goal stdev $\sigma_g$, prev stdev $\sigma_p$, EMA parameter $\alpha$
\State \textcolor{blue}{Phase I- Supervised Pre-Train Affordance Backbone}
\State Input Dataset $\mathcal{D}_v$ of ego-centric videos
\State Track $\mathcal{D}_v$ into $\mathcal{D_\xi}$ of trajectories
\State Convert $\mathcal{D_\xi}$ into $\mathcal{D}$ of (image, heatmap) pairs
\State Train $A(s)$ on $\mathcal{D}$ via supervised learning
\State \textcolor{blue}{Phase II- Online Control with Dynamic Planning}
\State $s \gets s_{start}$
\State $\textbf{p} \gets [1]_{i=1}^k$
\While{$s \neq s_{goal}$}
\State $\textbf{b}_{filtered} \leftarrow \texttt{Affordance\_Backbone}(s)$
\State $\textbf{g} \gets \left[ \mathcal{N}(x_i; g, \sigma_g) \right]_{i=1}^{k}$
\State $\textbf{v} \gets \textbf{b}_{filtered} * \textbf{g} * \textbf{p}$
\State $f_\pi \gets \argmax{(\textbf{v})}$
\State $\hat{a} \sim \pi_\lambda(s, f_\pi)$
\State $s \gets \text{Execute}(s, a)$
\State $\textbf{p} \gets \left[ \mathcal{N}(x_i; f_\pi, \sigma_p) \right]_{i=1}^{k}$
\EndWhile
\end{algorithmic}
\label{alg: lrn}
\end{algorithm}

\begin{algorithm}
\caption{\texttt{Affordance\_Backbone $A(s)$}}
\begin{algorithmic}
\Require EMA $\alpha$
\State $\text{heatmap} \gets \texttt{PredictHeatmap}(s)$
\State $\textbf{b} \gets \texttt{Project}(\text{heatmap})$
\State $\textbf{b}_{norm} \gets \textbf{b} / \sum_{i=1}^{k}b_i$
\State $\textbf{b}_{filtered} = \alpha * \textbf{b}_{norm} + (1-\alpha) * \textbf{b}_{filtered}$
\State \texttt{Return} $\textbf{b}_{filtered}$
\end{algorithmic}
\label{alg: affordance-module}
\end{algorithm}

\subsection{Learning Affordances from Unlabeled Videos}
\begin{figure*}[ht!]
    \centering
    \includegraphics[width=1.0\linewidth]{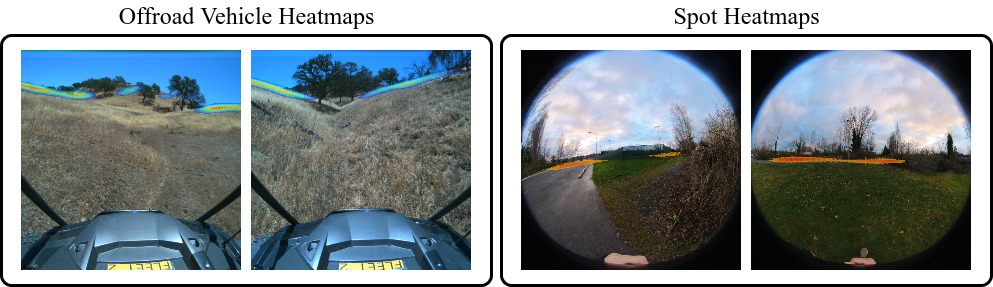}
    \caption{\textbf{Heatmaps computed by \texttt{LRN}.} The two images on the left show examples from the Racer Heavy. Blue is lower confidence and red is high confidence. \texttt{LRN} finds affordable spots between trees and on hill crests. The two images on the right show heatmap examples from the Insta360 camera that is mounted on the Spot robot. These are pictures from the human data collection. \texttt{LRN} correctly puts high score on the two forks in the road and to the side of bushes. For more qualitative results see Appendix~\ref{app:heatmaps}.}
    \label{fig:heatmap_example}
    \vspace{-10pt}
\end{figure*}

While we show results from learning image affordances from hand-labeled data (Fig.~\ref{fig:heatmap_example}), this approach proved to be tedious and scales poorly with more data. This raises the question - \emph{can we learn such affordances from unlabeled videos}? Concretely, we utilize unlabeled ego-centric videos to generate affordable image frontiers. The \textbf{key insight} here is that to generate labels analogous to human-annotation, the end of a trajectory should be representative of a good frontier from the starting state of this trajectory. To get access to the end position of the trajectory in the visual space from the starting state, if we have accurate access to the robot's 3D position across the trajectory then we can use the camera intrinsics to project the 3D points in the image space. However, accurate 3D is tricky in practice for long trajectories as small errors can result in points being in the sky or on lethal obstacles. Furthermore, this restricts us to using only robot data and ignoring abundant and easy-to-collect ego-centric videos - a person only needs to take a quick walk.

\begin{figure}[ht!]
    \centering
    \begin{subfigure}[t]{0.2\textwidth}
        \centering
        \includegraphics[width=\textwidth]{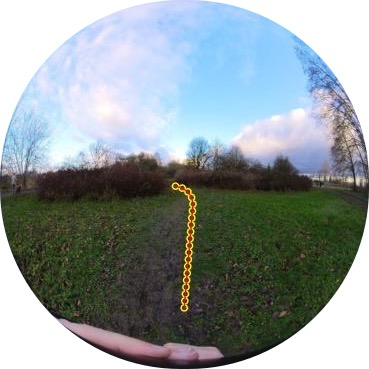}
        \caption{Human trajectory}
        \label{fig:image_b}
    \end{subfigure}%
    \begin{subfigure}[t]{0.2\textwidth}
        \centering
        \includegraphics[width=\textwidth]{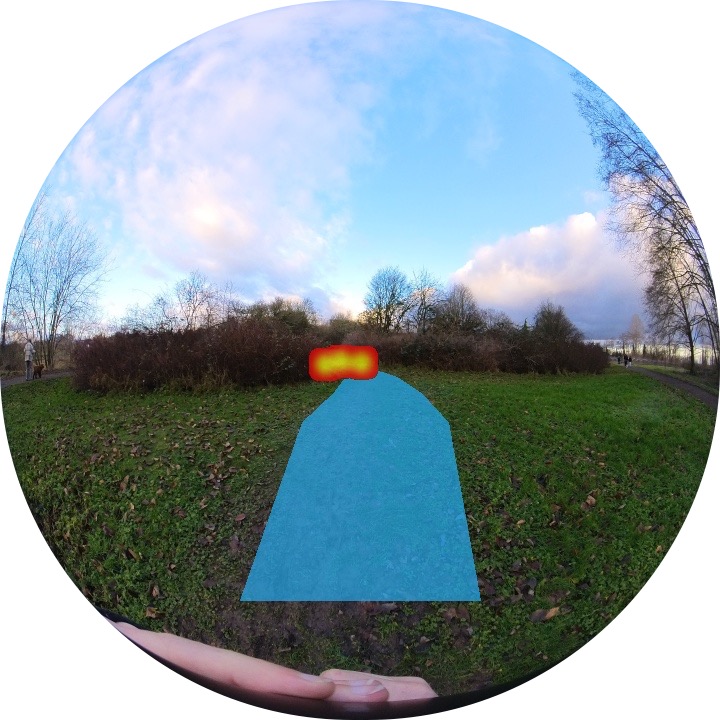}
        \caption{Hotspot visualization}
        \label{fig:image_c}
    \end{subfigure}
    \caption{\textbf{Learning Affordances from Unlabeled Videos.} The fisheye image~\ref{fig:image_b} shows the observation at the start of the trajectory and the path taken by the human. Fig.~\ref{fig:image_c} shows the hotspots computed by the automatic data labeling pipeline. The blue part is the path leading up to the hotspot and is labeled as 0. The end of the trajectory becomes the hotspot with yellow indicating high score of 1 and fades to red with a low score above 0.}
    \label{fig:auto_label}
    \vspace{-20pt}
\end{figure}

Since \texttt{LRN} only needs points in image space, a design choice we make is to forgo precise localization and instead use the video tracker model CoTracker \cite{karaev2023cotracker}. CoTracker tracks a grid of points in image space. To get the end of a trajectory, we run the video in reverse and select a subset of the grid right in front of the camera. Once the point becomes occluded (CoTracker provides this) we mark it as as affordable frontier. Fig.~\ref{fig:auto_label} shows the tracked trajectory on the left and a heatmap generated from the trajectory on the right. 

Once points are tracked, heatmaps are generated by taking the affordable frontier image points and marking them as 1 (affordable hotspot) and the rest of the trajectory as 0 (not affordable frontier). For other parts of the image we leave it unlabeled and do not incur a loss on predictions there. We found further marking the vertical column around the affordable hotspots as 0 reduced false positives, particularly in the sky.

As shown in Fig.~\ref{fig:bilevel} the affordance backbone uses the frozen image encoder to encode RGB camera images. We used SAM2~\cite{ravi2024sam2} for Racer Heavy tests and the lighter weight MobileSAM~\cite{mobile_sam} for Spot tests. We then train a small de-convolution decoder to predict heatmaps. We use an MSE loss between target and prediction to train the model with L2 regularization. Note, the Racer Heavy results, were obtained from an earlier version of \texttt{LRN} which was trained solely from human labeled images.

\subsection{Goal Conditioning}
Given the affordance heatmaps for each camera sensor, the scores are projected to a discrete set of angular bins using camera intrinsics. For each bin \texttt{LRN} takes the sum of scores falling in that bin for the given camera. Bins with scores less than a threshold $h_{thresh}$ are set to 0. The max is taken for overlapping bins between cameras, then the scores are normalized. Finally, a exponential moving average (EMA) filter is applied with a weight $\alpha$ to filter scores over time reducing fluctuations. If a vector of $k$ discretized bin scores is denoted as $\textbf{b}$, then
\begin{gather*}
\textbf{b} = [b_1, b_2,\dots,b_k]\\
\textbf{b}_{norm} = \frac{\textbf{b}}{\sum_{i=1}^{k}b_i}\\
\textbf{b}_{filtered} = \alpha * \textbf{b}_{norm} + (1-\alpha) * \textbf{b}_{filtered}
\end{gather*}
The goal conditioned cost function (Fig.~\ref{fig:bilevel}) takes in $\textbf{b}_{filtered}$ the goal angle $\mu_g$ and the previous selected heading $\mu_p$. The goal heading and the previously timestep's selected heading each define a Gaussian score centered on $\mu_g$ and $\mu_p$ previously. Similar to how the EMA filter reduces fluctuations in the individual scores, the previous selected heading is used to reduced fluctuations in the final selected heading. We apply these functions to the discrete bins to obtain a goal cost vector and a consistency cost vector for $k$ angular bins.
\begin{gather*}
\textbf{g} = \left[ \mathcal{N}(x_i; g, \sigma_g) \right]_{i=1}^{k}\\
\textbf{p} = \left[ \mathcal{N}(x_i; f_\pi, \sigma_p) \right]_{i=1}^{k}
\end{gather*}
Where, $\sigma_g$ and $\sigma_p$ are both fixed parameters. Finally all the vectors gets multiplied together to obtain final scores. The maximum scoring angle is then selected.
\begin{gather*}
\textbf{v} = \textbf{b}_{filtered}* \textbf{g} * \textbf{p}\\
f_\pi = \argmax{(\textbf{v})}
\end{gather*}
\section{Experimental Design}
\label{sec:experiments}
In this section, we instantiate \texttt{LRN} in two practical setups, namely, Spot, and a Racer Heavy platform both operating in outdoor environments. Our experiment design is motivated by empirically studying the following research questions:
\begin{itemize}
    \item \textbf{[Q1.]} Can the intermediate affordance representation proposed in \texttt{LRN} exhibit more efficient navigation capability as compared to other approaches? (See Sec.~\ref{sec:q1})
    \item \textbf{[Q2.]} Considering the connection between the quality of the affordance model against system performance, do better affordances lead to more efficient paths? (See Sec.~\ref{sec:q2})
    \item \textbf{[Q3.]} Does \texttt{LRN} generalize to out-of-distribution scenarios that were not seen during affordance model training? (See Sec.~\ref{sec:q3})
\end{itemize}

We now describe the details of local policies, \texttt{LRN} training, goal conditioned head, online evaluation courses, and, offline evaluation for both platforms. 

\subsection{Local Policies}
For both platform's local policies, we follow a traditional perception, planning, and control pipeline where perception creates a metric costmap, planning finds a path through it, and control roughly tracks that path re-planning as new information comes in.
\paragraph{Spot}
 We leverage Elevaton Mapping CuPy~\cite{miki2022elevation} a fast elevation mapping software. The local elevation map is a square with 16m width/height and created from a combination of depth cameras and an ouster OS1 lidar . The elevation map is converted to a costmap for planning by mapping slopes to cost via simple rules. We then use an $ARA^*$ planner~\cite{likhachev2003} over a lattice to plan a path through that map. A carrot point 3m ahead along the planned path is used to compute a body frame velocity which is passed to Spot's internal navigation system. The internal navigation system handles locomotion and performs some obstacle avoidance. 

The entire stack and \texttt{LRN} is run on a Jetson Orin AGX in realtime. To achieve real time performance \texttt{LRN} uses a MobileSAM image encoder~\cite{mobile_sam} and achieves a ~4hz inference with autonomy also running. At runtime \texttt{LRN} performs inference on both raw front and rear fisheye lens from a Insta360 camera. Once a heading is computed it is sent to the planner as a goal just outside the range of the costmap in the direction of the heading. 

\paragraph{Racer Heavy}
We had an opportunity to deploy on a Racer Heavy platform. It is a 12 ton tracked vehicle  equipped with three front facing cameras and one rear camera amongst other lidar and odometry sensors. The local stack in this demonstration was a heavily optimized perception planning and control stack with a circular costmap of radius 50m. The planner is a search-based planner that will plan to the goal but stop once it reaches a frontier node at the edge of the costmap within a short tolerance of the goal heading. This allows the planner some flexibility in case an obstacle is blocking the exact goal. \texttt{LRN} was run on all four time-synced camera images. We opted for a larger SAM2~\cite{ravi2024sam2} image encoder which ran at ~4hz while the autonomy stack was also running.
\begin{figure*}[ht!]
    \centering
    \includegraphics[width=1.0\linewidth]{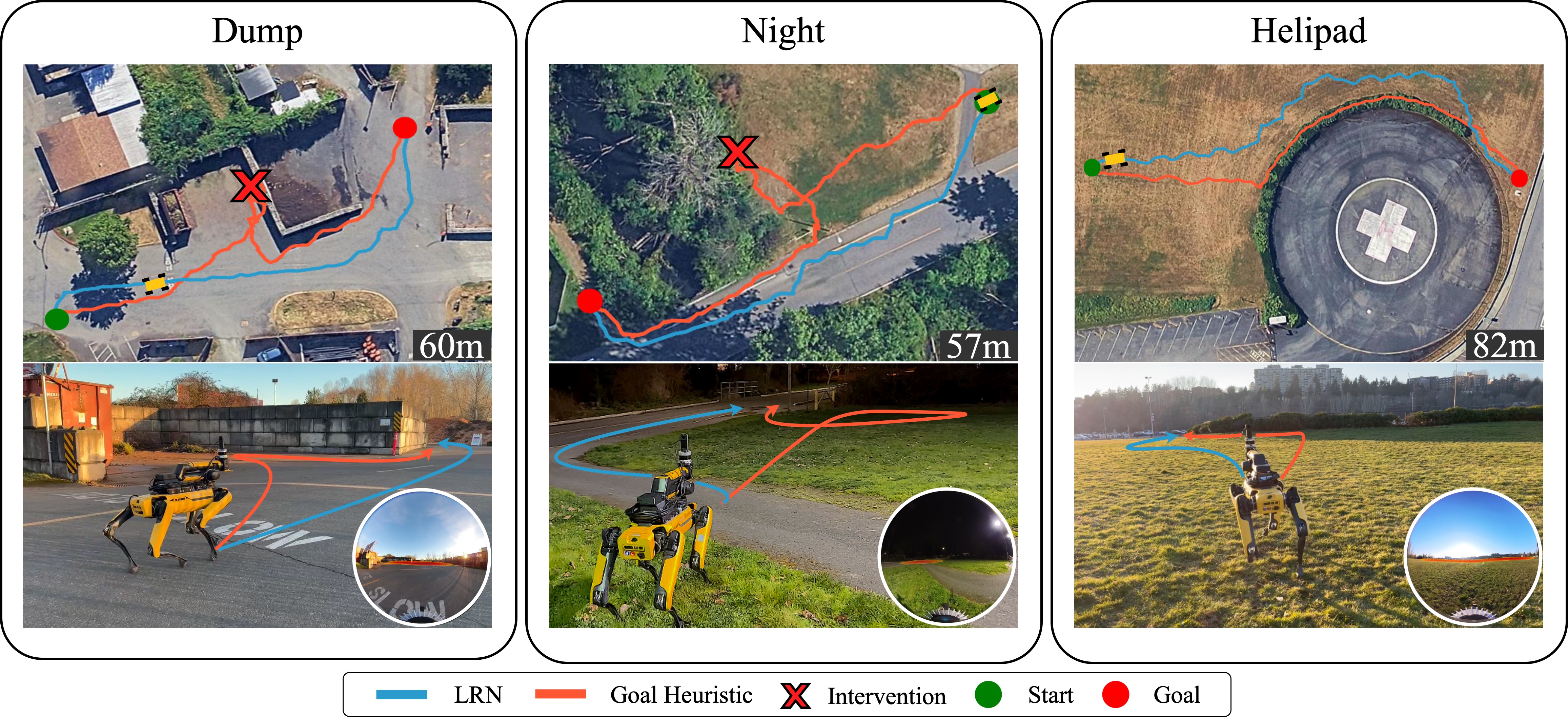}
    \caption{\textbf{GPS plots of \texttt{LRN} and Goal Heuristic Baseline on each course.} Where the Goal Heuristic blindly charges towards the goal, the \texttt{LRN} makes earlier decisions to avoid difficult terrain. The round image at the bottom right of each Spot image shows the image the robot observes and the corresponding heatmaps. Red denotes high scores. In the dump and night scenario, there is high score to the side of the wall and on the sidewalk to the bridge. For the helipad course, the \texttt{LRN} incorrectly puts some heat on the bushes also, highlighting some sub-optimal \texttt{LRN} predictions. For qualitative results of Trav. Depth and NoMaD see Appendix~\ref{app:spot_gps}.}
    \label{fig: spot_qual}
    \vspace{-15pt}
\end{figure*}
\subsection{\texttt{LRN} Training}
\paragraph{Spot}
We collected datasets in two semi-urban environments by walking around with an Insta360 camera collecting videos totaling 54 minutes. Since the points of interest are the ground, the tracked points can drift over long times. In order to have clean data, we chopped the videos into 2 minute segments. The videos were then processed by our automatic labeling pipeline to produce a dataset of 92,711 heatmap labels. The training used an MSE loss with L2 weight regularization to train a 5 layer decoder network.

\paragraph{Racer Heavy}
The Racer Heavy experiment used an early version of \texttt{LRN} trained on human labeled data. The dataset was 1,901 human labeled heatmap images from a California oak savanna. Notably the labeled images were only front-facing camera images from a different Racer Heavy than the one we deployed on. We asked humans to label all affordable frontiers in each image. They additionally had access to future/past observations and side-cameras for context (some example images are shown in the supplementary materials). Human selected regions were positive labels and the rest of the image is assumed negative. We additionally add some Gaussian blur around positive labels as we found the smooth transitions helped with training. To improve robustness to visual variations, we further augmented the dataset by applying random color jitter, sharpness adjustments, rotations, and blur increasing the dataset size to 11,406 images. Our loss was a pixel-wise MSE loss with additional L2 regularization on the weights. 

\subsection{Goal Conditioned Head}
\paragraph{Spot}
For the goal-conditioned level we use an angle discretization of 5 degree width bins, $h_{thresh}=0.7$, $\alpha=0.1$, $\sigma_g=90$, and $\sigma_p=110$. Additionally, to avoid walking past the goal we implement two additions. First, when the robot is within 30m of the goal it linearly reduces the $\sigma_g$ based on its distance to goal. Second, when the robot is within 12m of the goal it switches to heading straight to the goal.
\paragraph{Racer Heavy}
For Racer Heavy we used a threshold of $h_{thresh} = 0.15$. The EMA filter used to reduce noise in affordance scores was set to $\alpha=0.1$. For goal costs we used $\sigma_g = 70$ and $\sigma_p=100$ weighting previous predictions less than the goal direction. When the robot got within 75m of the goal it would return to heading straight for the goal. There was no linear decrease in $\sigma_g$ like in the Spot experiments.

\subsection{Real World Robot Evaluation}
\label{sec:real_world}
For real robot experiments we compare to a set of baselines described below.
\begin{itemize}
\itemsep-0.1em
    \item \textbf{Goal Heuristic} is a simple baseline that plans the shortest path to a set of points at the edge of the costmap nearest the goal. Goal Heuristic is implemented using the Goal Conditioned Head of LRN but by providing a uniform distribution for $b_{filtered}$ so as to focus comparison on the image affordances from LRN.
    \item \textbf{NoMaD}~\cite{sridhar2023} is SOTA visual exploration approach which we fine-tuned on a dataset collected on robot in the same type of environment as our test sites. It uses a diffusion policy that predicts trajectories given a robot's ego camera image and a target image. We still leverage Spot's internal obstacle avoidance but forgo the local perception and planning as NoMAD is a full end-to-end system. NoMAD has been shown to work well with a topological map of images. The offroad driving setting is different in that the goal context is a waypoint. The experiments with NoMAD also highlight the potential issues of long horizon navigation in offroad environments with a sparse topological map.
    \item \textbf{Traversability + Depth Anything V2} combines a visual traversability estimator trained on our dataset and Depth Anything V2 monocular depth estimator. The intuition is that distant traversable points should be hotspots. We combine the two outputs by normalizing their scores and multiplying them to produce a heatmap similar to LRN, which are then used instead of the LRN hotspots. For Racer Heavy, we choose the V-Strong~\cite{jung2024} traversability model which had been trained for the California hills environment. For Spot, a V-Strong model was not available so we trained a traversability model using the same model and training as \texttt{LRN} but instead of considering only the hotspot to be 1 in the loss, we mark the whole trajectory as it is traversable. To improve traversability further we take a trick from V-Strong and expand the traversable region by making a SAM mask seeded from the robot's path. See Appendix~\ref{app:travdepth} for more details and visualizations of these components.
\end{itemize}
For the Spot tests we compared against all baselines and for Racer Heavy we compared against Goal Heuristic. For LRN, all test sites were unseen during training but from similar environments to the training set.

\paragraph{Spot}
To illustrate the challenges of using a limited range metric map, we found specific scenarios in semi-urban environments that showcase the failures practitioners would encounter with a Goal Heuristic. We tested on three courses: dump, night, and helipad as seen in Fig.~\ref{fig: spot_qual}.

\begin{itemize}
    \item \textbf{Dump:} This site has a big wall (similar to a classic bug trap) and the goal is behind it 60m from the robot's start position. The local system has good perception within its local map but the wall is outside the range of the local system until it gets close.

    \item \textbf{Night:} This night course is partially lit by overhead street lights. The robot must cross a bridge and get to a point on the other side of the river 57m away. The straight line between the start and the goal is blocked by a wall of bushes and the river. Further left is a wide bridge that crosses the river.
    \item \textbf{Helipad:} In this course, the robot must get to the other side of the helipad 82m away. The environment is an open field except the straight line between the robot and the goal is blocked by a wall of bushes. 
\end{itemize}
\begin{figure*}[ht!]
    \centering
    \includegraphics[width=1.0\linewidth]{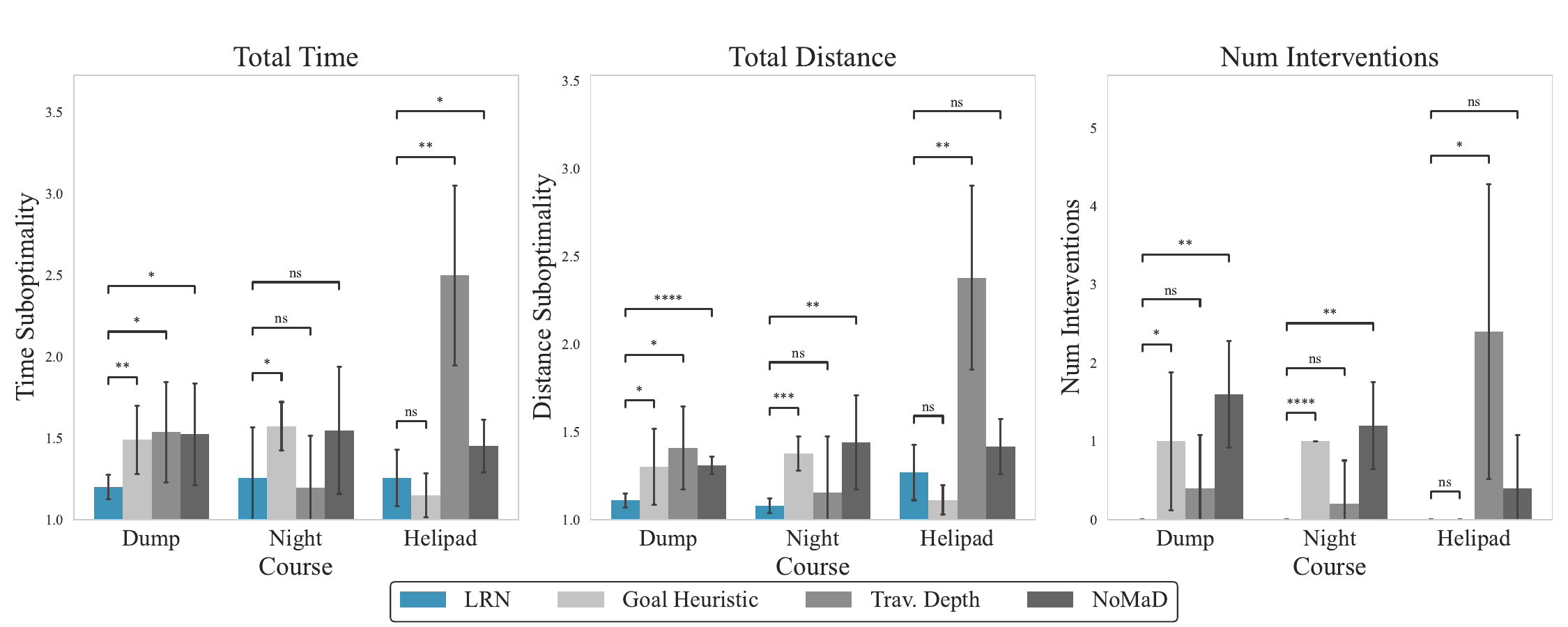}
    \caption{\textbf{Comparison of \texttt{LRN}, Goal Heuristic, Trav. Depth, and NoMaD on Spot tests.} We report average and 95 \% confidence intervals for 5 real world experiments. Time and Distance suboptimality are with respect to human baseline runs. We use asterisks $^*$ to denote statistical significance : $^* p < 0.05$, $^{**} p < 0.01$, $^{***} p < 0.001$, and $^{****} p < 0.0001$}
    \label{fig: spot_metrics}
    \vspace{-15pt}
\end{figure*}
Each approach was tested for five trials per course due to limited testing windows given weather and human traffic constraints. We evaluate the performance via Total Distance, Human Interventions, and Total Time. Human interventions were taken when the system was not making progress or the robot was entering a dangerous situation.  Interventions were performed by facing the robot towards the goal and seeing if it will head in a reasonable direction autonomously. If not, we would walk it towards the goal until we saw a reasonable navigation plan. As the focus is on long-range navigation, we did not count interventions where local perception failed to perceive an obstacle.

\paragraph{Racer Heavy}
The Racer Heavy test was intended as a more holistic test of \texttt{LRN} on a full system. Thus, we provide each approach with a single waypoint 660m away which crosses three hills two of which have dense clusters of trees that should be avoided. We use similar metrics and guidelines for human interventions as Spot. Given time constraints, we were only able to run each method for one trial demonstrating the system but not fully testing it.

\subsection{Offline Evaluation}
To evaluate the quality of affordance heatmaps learned offline we perform an evaluation on a human labeled test sets of 330 images for Spot and 315 images for Racer Heavy unseen during training but from the same environments. Since no other affordance heatmap predictor exists to our knowledge we compare against Traversability + Depth Anything V2~\cite{yang2024}. 

To evaluate these methods we binarize the target heatmaps with a threshold of 0.15 which makes everything that is hot in Fig.~\ref{fig:auto_label} 1 and everything else 0. We use Area Under the Reciever Operator Curve (AUROC), F1 score, Precision, Recall, False Positive Rate (FPR), and False Negative Rate (FNR). All metrics are [0, 1].

\section{Results}
\label{sec:results}
We perform a total of 5 tests per approach on Spot for a total of 60 trials. For ablation, we perform an additional 30 tests at the dump test site with 5 trials per heatmap threshold.

\subsection{\textbf{[Q1.]} Can the intermediate affordance representation proposed in \texttt{LRN} exhibit more efficient navigation capability as compared to other approaches?}
\label{sec:q1}
We first investigate the efficacy of \texttt{LRN} when compared to the baselines described in \ref{sec:real_world}. For Spot, we report in Fig.~\ref{fig: spot_metrics} that \texttt{LRN} outperformed Goal Heuristic on all metrics on Dump and Night courses and was comparable on Helipad course. Compared to Trav. Depth and NoMaD \texttt{LRN} outperformed on Dump and Heli mostly and saw competitive performance on Night course. \texttt{LRN}, Trav. Depth and NoMaD see a higher total distance in Helipad compared to the Goal Heuristic due to these predictive models switching directions more frequently causing wandering. Finally, we see no interventions with \texttt{LRN} on any course whereas all other methods incur some human interventions.

Qualitative analysis shows in Fig.~\ref{fig: spot_qual} that \textbf{\texttt{LRN} can make earlier decisions to avoid large obstacles} compared to Goal Heuristic highlighting its longer range reasoning ability (See Appendix~\ref{app:spot_gps} for more qualitative results). The GPS paths for \texttt{LRN} visibly are more jagged, particularly in helipad due to some switching of \texttt{LRN} directions which slows the robot down.

\begin{figure}[t!]
    \centering
    \includegraphics[width=1.0\linewidth]{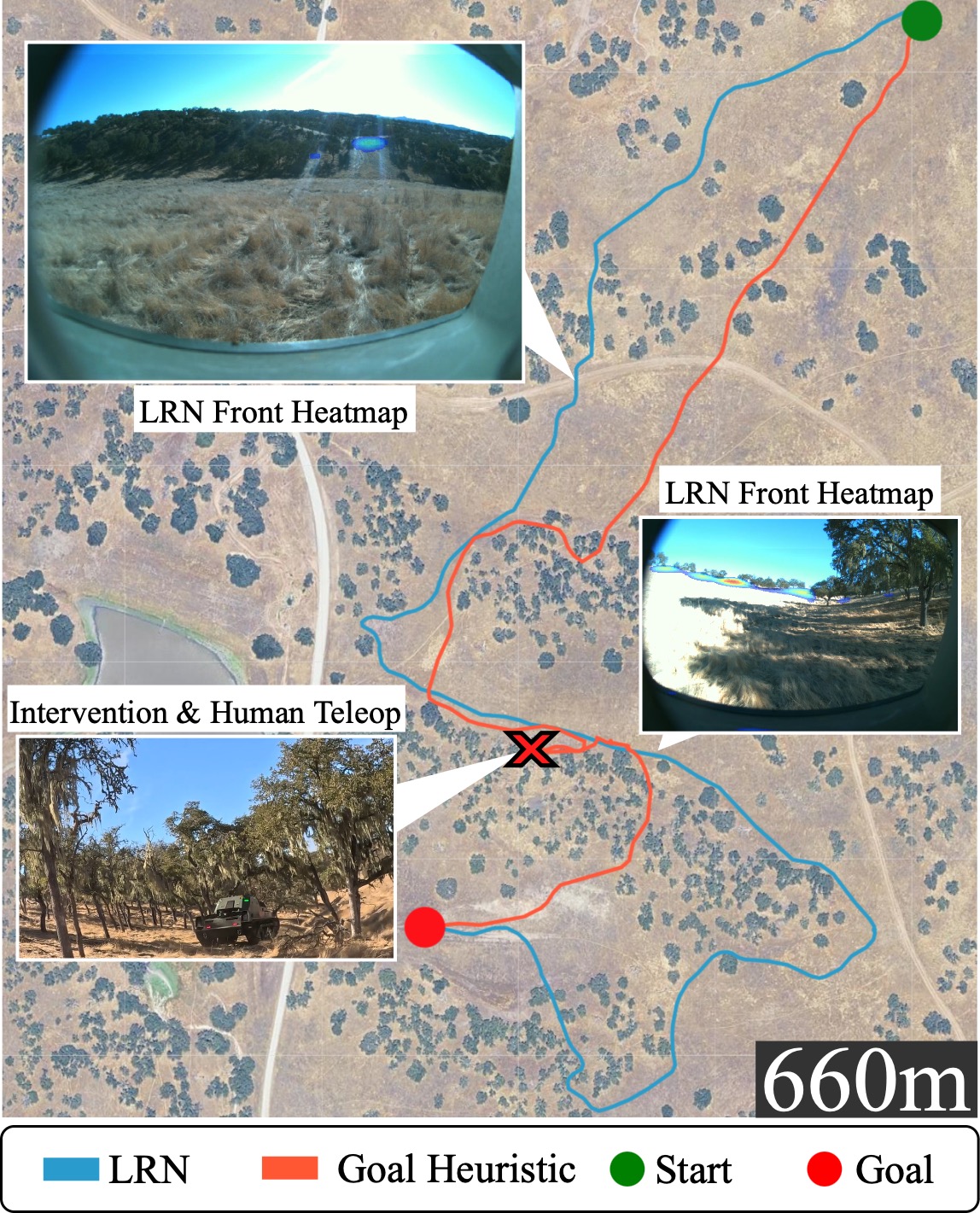}
    \caption{\textbf{Full scale \textbf{LRN} and Goal Heuristic demonstration}. Paths taken by baseline Goal Heuristic and \texttt{LRN} systems given the same start and goal. The intervention and human teleop for the Goal Heuristic was due to it pushing into dense trees near a ditch. The \texttt{LRN} avoids the dense forests on all the hills. It  misses an opening towards the end and takes a slightly longer path to the goal.}
    \label{fig: e6_lrn_path}
\end{figure}

For the Racer Heavy, we report in Table.~\ref{table: racer_metrics} that \textbf{\texttt{LRN} achieves a higher average and max speed while having zero interventions}. This is likely achieved by it taking a longer, more open route around dense and difficult terrain. While this test was only of sample size one, we think this shows promise in \texttt{LRN}'s ability to work in real navigation tasks with a full system. 
Qualitative analysis shown in Fig.~\ref{fig: e6_lrn_path} shows the Goal Heuristic baseline (orange) heads straight into the treeline on the second hill. Luckily it backs out and finds a path, only to get stuck on the next hill trapping itself. This required a 60m human intervention from which it completed the course. In contrast, \texttt{LRN} (blue) found a clear opening on the second hill, avoiding the tree line. On the third hill, it opted to circumvent the entire dangerous tree line. Once past the treeline, it started heading to the goal but found greater heat Southward veering it off course but eventually finding the goal without intervention. This missed turn is likely due to our fixed $\sigma_g$ which weighted the goal the same no matter how close the robot is.

Further, we report offline test results on tests sets shown in Table~\ref{table: offline_metrics}. We observe that \texttt{LRN} outperforms Trav. Depth for all metrics. We note that the performance for both Racer Heavy and Spot is better in the human labeled data. This is not surprising as any error in approximating the correct affordances induces a bias thereby impacting performance. When trained with human-expert labeled data, \texttt{LRN} can be viewed as equipped with oracle affordance labels, resulting in better performance. 

\textbf{Remark I:} Through these experiments, we see \texttt{LRN} drives down interventions and tends to take shorter paths by seeing and avoiding dangerous terrain early suggesting it can improve overall navigation performance. 

\begin{table}[t!]
    \resizebox{\linewidth}{!}{
    \setlength{\tabcolsep}{2pt} %
    \begin{tabular}{lcccc}
    \toprule
        Algorithm  & Interventions & Avg./Max Speed (m/s)& Distance (m) & Time (s) \\ \midrule
        Goal Heuristic & 1 & 4.02 / 8.09 & \textbf{941.79} & \textbf{247.00}\\
        \texttt{LRN} & \textbf{0} & \textbf{4.98 / 8.66}  & 1435.09 & 289.00 \\ 
        \bottomrule
    \end{tabular}
    }
    \caption{\textbf{Run metrics for the Racer Heavy demonstration} shown in Fig.~\ref{fig: e6_lrn_path}.}
    \label{table: racer_metrics}
    \vspace{-20pt}
\end{table}
\textbf{Remark II:} We note that \texttt{LRN}'s performance is dependent on the accuracy of the affordance model. Having access to perfect affordances e.g. in human-expert labels used for training in Racer Heavy induces lesser bias and therefore better performance. In contrast, automatic labeling of affordance labels while less burdensome  can result in less accurate affordances, and therefore induce higher bias. We see this performance gap in offline evaluation shown in Table~\ref{table: offline_metrics}. In Spot Results, \texttt{LRN} sometimes exhibits switching and wandering behavior chasing various affordable frontiers. This is in part due to not modeling the correct affordances. We note wandering can also be the cause of improper tuning of the goal conditioned head or the goal conditioned head not fully capturing the true goal-conditioned cost.

\begin{table}[ht!]
\centering
\resizebox{0.95\columnwidth}{!}{ %
\begin{tabular}{@{}llcccc@{}}
\toprule
\textbf{System}   & \textbf{Metric}        & \textbf{\texttt{\texttt{LRN Auto}}} & \textbf{\texttt{\texttt{LRN Hand}}}                           & \textbf{Trav. Depth}\\ \midrule
Racer Heavy   & AUROC $\uparrow$        & 0.63                & \textbf{0.84}       & 0.56\\
                  & F1 $\uparrow$           & 0.11                & \textbf{0.52}       & 0.09\\
                  & Prec. $\uparrow$        & 0.08                & \textbf{0.51}       & 0.07\\
                  & Rec. $\uparrow$         & 0.17                & \textbf{0.52}       & 0.13\\
                  & FPR $\downarrow$        & 0.03                & \textbf{0.01}       & 0.03\\ 
                  & FNR $\downarrow$        & 0.83                & \textbf{0.48}       & 0.87\\
                \\ \midrule
Spot              & AUROC $\uparrow$        & \textbf{0.93}       & 0.77                & 0.61\\
                  & F1 $\uparrow$           & 0.10                & \textbf{0.32}       & 0.14\\
                  & Prec. $\uparrow$        & 0.06                & \textbf{0.30}       & 0.14\\
                  & Rec. $\uparrow$         & \textbf{0.35}       & \textbf{0.35}       & 0.13\\
                  & FPR $\downarrow$        & 0.01                & \textbf{0.0}        & \textbf{0.0}\\ 
                  & FNR $\downarrow$        & \textbf{0.65}       & \textbf{0.65}       & 0.87\\
\end{tabular}
}

\caption{\textbf{Classification metrics for heatmap backbone on test set.} Prec. and Rec. stand for Precision and Recall. The F1/Prec/Rec/FPR/FNR/ are from the highest scoring heatmap thresholds for each method: Trav. Depth and \texttt{LRN}.}
\label{table: offline_metrics}
\vspace{-15pt}
\end{table}
\subsection{\textbf{[Q2.]} Considering the connection between the quality of the affordance model against system performance, do better affordances lead to more efficient paths?}
\label{sec:q2}

\begin{figure}[ht!]
    \centering
    \includegraphics[width=\linewidth]{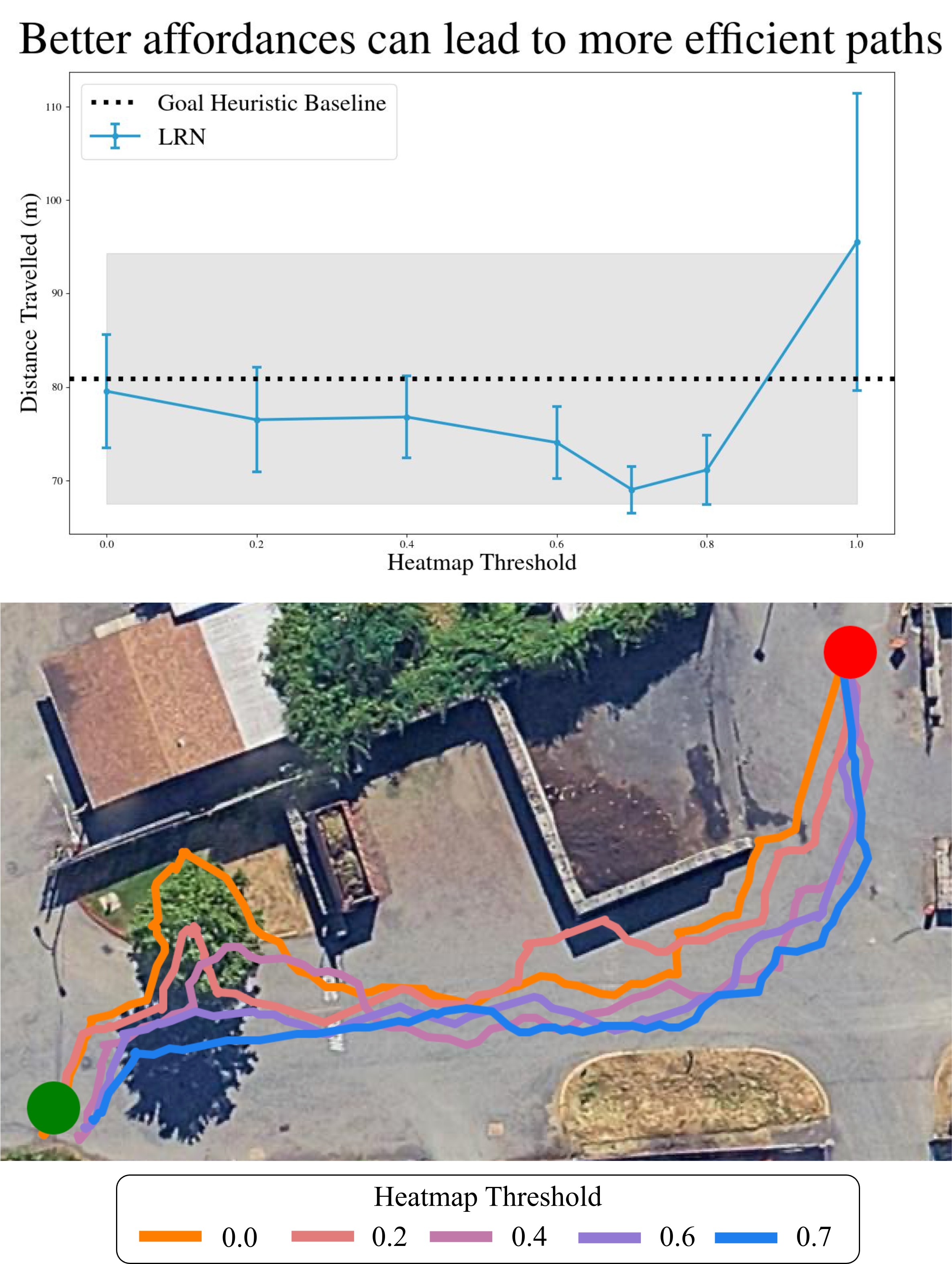}
    \caption{\textbf{Better affordances can lead to more efficient paths} To investigate this claim, we modify the heatmap threshold affecting the affordance set size. Low threshold means \texttt{LRN} considers more potentially poor options i.e. everything is affordable. High threshold means \texttt{LRN} sees very few or no options i.e. nothing is affordable. We note that an intermediate value of affordance threshold (i.e. $0.7$) gives the optimal affordance heatmaps resulting in \texttt{LRN} to take shorter travel distance, and drives most gains when compared to the Goal Heuristic. Our optimal threshold is $0.7$. We fit a line from threshold $0.0$ to $0.7$ and test the hypothesis that the slope is less than 0. We find there is a correlation ($p < 0.05$) between affordance quality and traversal distance.}
    \label{fig: ablation}
    \vspace{-20pt}
\end{figure}

To evaluate the connection between affordance quality and navigation performance, we perform an ablation on the Dump course as shown in Fig.~\ref{fig: ablation}. To vary the affordance heatmaps quality, we adjust the heatmap threshold $h_{thresh}$ in range 0 to 1.0. At 0 much of the environment is predicted as affordable causing the robot to take a more direct path to the goal and getting stuck in more local minima like the Goal Heuristic. At 1.0 almost none of the environment is deemed affordable and the robot switches between heading directly towards the goal and a random hotspot that appear infrequently. Between those extremes, we see how a $h_{thresh}$ best tuned for the system directly translates to the best navigation performance.

\textbf{Remark III:} By adjusting the heatmap quality via $h_{thresh}$ we see a correlation between improved affordances and improved navigation performance, indicating that there is an intermediate affordance value of affordance set size, that drives most gains when compared to the Goal Heuristic. Considering almost all directions as affordable or few affordable both can hurt the system which is intuitive. But we note, better affordances may not always lead to shorter paths as \texttt{LRN} is a heuristic that cannot predict what the environment will be like beyond view.

\subsection{\textbf{[Q3.]} Does \texttt{LRN} generalize to out-of-distribution scenarios that were not seen during affordance model training?}
\label{sec:q3}

Next, we investigated \texttt{LRN}'s ability to generalize in out-of-distribution scenarios. Each robot's LRN model was trained on data similar to the test environment excluding the test courses. LRN showed generalization in embodiment: for Racer Heavy it was trained on a different vehicle's camera data and for Spot it was trained from human walking videos. Spot further generalized to nighttime scenarios performing well (Fig.~\ref{fig: spot_metrics}) despite being trained only on daytime data, due to robust visual embeddings from MobileSam.

\textbf{Remark IV:} We reported multiple cases of generalization from \texttt{LRN} answering this question in the affirmative. We attribute this mostly to learning goal agnostic intermediate affordance representation from strong visual SAM embeddings. We note LRN inherently being a supervised learning method is most robust to scenes close to its training distribution.

\section{Related Work}
\label{sec:related_work}
\begin{itemize}
    \item \textbf{Subgoal Planning}(\cite{stein2018,gao2023, yamauchi1997, qi2020}): These works focus on planning to subgoals or frontiers and \texttt{LRN} would fall into this category. Most similiar to our work is ~\citet{stein2018} which learns whether a subgoal will reach the goal and the cost to go. This work focuses on indoor environments and uses lidar input whereas our work focus on outdoor environments with camera input which does not have depth information. A method using camera's has also been explored~\cite{qi2020}. They learn where the robot can and can't go by backprojecting where the robot went into the image space. While similar in design they focus on simulated video game worlds with perfect depth to enable accurate projection.
    \item \textbf{Learning From Videos}(\cite{kumar2019learningnavigationsubroutinesegocentric}): Prior works demonstrate learning navigation subroutines from ego-centric videos. However, these are more local and short horizon subroutines like turn-left or go-through-door and do not reason about unknown space far away.
    \item \textbf{Reasoning about Unknown Space}(\cite{bachrach2013}) \citet{bachrach2013} focuses on the same problem but takes an entirely different approach formulating the it as a state estimation problem over trajectories. They learns to estimate which trajectory bundles are viable in simulation using camera input. Trajectory bundles are sets of trajectories, and a viable one is one where there exists one trajectory that is collision free.
    \item \textbf{Near to Far learning}(\cite{hadsell2009, bajracharya2008, wang2010, moghadam2010, nava2019, zhang2018}): These works try to extend the metric map at a lower resolution. While these methods have shown to extend the map some, they still require depth for projecting to a map (or make a flatground assumption) which has limited maximum range and can be further limited by occlusions. 
    \item \textbf{Visual navigation}(\cite{shah2023, sridhar2023, meng2019}): These works are very similar in practice but target local navigation. They require topological maps for longer range navigation. We compare against NoMaD~\cite{sridhar2023} on the shorter Spot tests. Further, LRN could be combined with these local methods to enable longer range reasoning when a map is not available. 
    \item \textbf{Traversability Prediction}(\cite{mattamala24, schmid2022, jung2024, wellhausen2020}): These works predict traversability in image space. While this is a similar task to ours, traversability alone is not sufficient to find affordable frontiers. We show this via comparing with the Traversability + Depth Anything V2 baseline.
    \item \textbf{Cost Inpainting}(\cite{shaban2022, fahnestock2024, meng2023}): These methods attempt to inpaint cost in unknown space. Notably~\cite{fahnestock2024} uses a diffusion model to predict a larger map given the local costmap, but does not use sensor information besides the costmap for prediction.
\end{itemize} %
\section{Limitations} 
\label{sec:limitations}
While \texttt{LRN} has shown better overall long range navigation, our method is not without its limitations. We here discuss key failure modes. 

First, LRN does not reason about depth explicitly. Without depth, we are implicitly assuming that the angular distance to goal from an LRN hotspot is a sufficient proxy for distance to goal. This assumption can break when two or more hotspots are equal angular distance from the goal heading but in reality, one is much closer to the goal. This appears as occasional wandering on the Helipad and Racer Heavy courses due to open environments with many hotspots in LRN predictions. That said, LRN recovered by eventually finding more direct hotspots and reaching the goal without intervention. Beyond incorporating depth, this issue can be addressed by adding a detector for wandering behavior and reducing $\sigma_p$ encouraging sticking to one decision or reverting to Goal Heuristic until fluctuations in LRN stabilize. A good signal for this is non-monotonic erratic fluctuations is distance to goal, which may not always be decreasing (e.g. going further around an obstacle), but should change smoothly.

Second, \texttt{LRN} can exhibit switching behavior due to fluctuating heat scores. Small fluctuations in score between two very different directions causes the robot to switch back and forth. This problem motivated the EMA filter and previous heading gaussian score but we found it does not completely alleviate the issue and more exploration is needed. We would like to explore learning the goal conditioned head with history to see if it can learn to maintain consistent headings.

Third, while some heatmaps seemed reasonable in online tests, we noticed more optimism from the Spot model where it puts heat on obstacles near an opening. We attribute this to the automated labels which some have small tracking errors putting heat on the edges of openings. Future work on reducing tracking error or filtering bad labels could improve performance on this front.

Finally, \texttt{LRN} is a heuristic for exploring unknown space. While we show it can be an improved heuristic over other methods, all heuristic frontier approaches suffer from not truly knowing the whole environment. Thus \texttt{LRN} cannot guarantee improved performance because a frontier that looks good from one perspective may actually lead down an unseen bad path. Incorporating history into the LRN could help so if the robot reached a dead end it could remember a previous hotspot and backtrack to that position. %
\section{Conclusion} 
\label{sec:conclusion}
\textbf{In summary,} we presented \texttt{LRN}, a novel method for thinking beyond metric costmaps to make less myopic navigation decisions, by leveraging an intermediate affordance representation from solely video data on top of a local navigation system or policy. Through extensive experiments, we demonstrated that better affordances can lead to better performance by making less myopic decisions, that \texttt{LRN} exhibits generalization to OOD data, and reported overall improved long-range navigation as compared to multiple baselines both qualitatively and quantitatively, with tests on two very different platforms.

\textbf{Future work} requires considering how sparse depth can improve performance when available. A key open question is how to incorporate memory of previous affordable frontiers into navigation decisions alongside incorporating history of observations to better handle dead-ends and wandering.

\section{Acknowledgments}
This research was developed with funding from the Defense Advanced Research Projects Agency (DARPA). 
\bibliographystyle{plainnat}
\bibliography{references}

\begin{thebibliography}{30}
\providecommand{\natexlab}[1]{#1}
\providecommand{\url}[1]{\texttt{#1}}
\expandafter\ifx\csname urlstyle\endcsname\relax
  \providecommand{\doi}[1]{doi: #1}\else
  \providecommand{\doi}{doi: \begingroup \urlstyle{rm}\Url}\fi

\bibitem[Bachrach(2013)]{bachrach2013}
Abraham Bachrach.
\newblock \emph{Trajectory bundle estimation For perception-driven planning}.
\newblock Phd thesis, Massachusetts Institute of Technology (MIT), 2013.

\bibitem[Bajracharya et~al.(2008)Bajracharya, Tang, Howard, Turmon, and
  Matthies]{bajracharya2008}
Max Bajracharya, Benyang Tang, Andrew Howard, Michael Turmon, and Larry
  Matthies.
\newblock Learning long-range terrain classification for autonomous navigation.
\newblock In \emph{{IEEE} International Conference on Robotics and Automation},
  pages 4018--4024, 2008.
\newblock \doi{10.1109/ROBOT.2008.4543828}.

\bibitem[Cormen et~al.(2009)Cormen, Leiserson, Rivest, and Stein]{cormen2009}
Thomas~H. Cormen, Charles~E. Leiserson, Ronald~L. Rivest, and Clifford Stein.
\newblock \emph{Introduction to Algorithms, Third Edition}.
\newblock The MIT Press, 3rd edition, 2009.
\newblock ISBN 0262033844.

\bibitem[Fahnestock et~al.(2024)Fahnestock, Fuentes, Osteen, Ancha, and
  Roy]{fahnestock2024}
Ethan Fahnestock, Erick Fuentes, Philip~R Osteen, Siddharth Ancha, and Nicholas
  Roy.
\newblock Learning semantic traversability priors using diffusion models for
  uncertainty-aware global path planning.
\newblock In \emph{{IEEE} International Conference on Robotics and Automation},
  2024.

\bibitem[Frey et~al.(2024)Frey, Mattamala, Piotr, Chebrolu, Cadena, Martius,
  Hutter, and Fallon]{mattamala24}
Jonas Frey, Matias Mattamala, Libera Piotr, Nived Chebrolu, Cesar Cadena, Georg
  Martius, Marco Hutter, and Maurice Fallon.
\newblock Wild visual navigation: Fast traversability learning via pre-trained
  models and online self-supervision.
\newblock In \emph{Robotics: Science and Systems}, 2024.

\bibitem[Gao et~al.(2023)Gao, Wu, Yang, and Ji]{gao2023}
Yan Gao, Jing Wu, Xintong Yang, and Ze~Ji.
\newblock Efficient hierarchical reinforcement learning for mapless navigation
  with predictive neighbouring space scoring.
\newblock \emph{IEEE Transactions on Automation Science and Engineering}, 2023.
\newblock \doi{10.1109/TASE.2023.3312237}.

\bibitem[Hadsell et~al.(2009)Hadsell, Sermanet, Ben, Erkan, Scoffier,
  Kavukcuoglu, Muller, and LeCun]{hadsell2009}
Raia Hadsell, Pierre Sermanet, Jan Ben, Ayse Erkan, Marco Scoffier, Koray
  Kavukcuoglu, Urs Muller, and Yann LeCun.
\newblock Learning long-range vision for autonomous off-road driving.
\newblock \emph{Journal of Field Robotics}, 26\penalty0 (2):\penalty0 120--144,
  2009.
\newblock \doi{https://doi.org/10.1002/rob.20276}.

\bibitem[Jung et~al.(2024)Jung, Lee, Meng, Boots, and Lambert]{jung2024}
Sanghun Jung, JoonHo Lee, Xiangyun Meng, Byron Boots, and Alexander Lambert.
\newblock V-strong: Visual self-supervised traversability learning for off-road
  navigation.
\newblock \emph{icra}, 2024.

\bibitem[Karaev et~al.(2024)Karaev, Rocco, Graham, Neverova, Vedaldi, and
  Rupprecht]{karaev2023cotracker}
Nikita Karaev, Ignacio Rocco, Benjamin Graham, Natalia Neverova, Andrea
  Vedaldi, and Christian Rupprecht.
\newblock {CoTracker}: It is better to track together.
\newblock \emph{ECCV}, 2024.

\bibitem[Kumar et~al.(2019)Kumar, Gupta, and
  Malik]{kumar2019learningnavigationsubroutinesegocentric}
Ashish Kumar, Saurabh Gupta, and Jitendra Malik.
\newblock Learning navigation subroutines from egocentric videos.
\newblock \emph{Proceedings in Machine Learning Research}, 2019.

\bibitem[Likhachev et~al.(2003)Likhachev, Gordon, and Thrun]{likhachev2003}
Maxim Likhachev, Geoffrey~J Gordon, and Sebastian Thrun.
\newblock $ara^*$ : Anytime $a^*$ with provable bounds on sub-optimality.
\newblock In \emph{Advances in Neural Information Processing Systems}, 2003.

\bibitem[Meng et~al.(2019)Meng, Ratliff, Xiang, and Fox]{meng2019}
Xiangyun Meng, Nathan Ratliff, Yu~Xiang, and Dieter Fox.
\newblock Neural autonomous navigation with riemannian motion policy.
\newblock \emph{CoRR}, 2019.

\bibitem[Meng et~al.(2023)Meng, Hatch, Lambert, Li, Wagener, Schmittle, Lee,
  Yuan, Chen, Deng, Okopal, Fox, Boots, and Shaban]{meng2023}
Xiangyun Meng, Nathan Hatch, Alexander Lambert, Anqi Li, Nolan Wagener, Matt
  Schmittle, JoonHo Lee, Wentao Yuan, Zoey Chen, Samuel Deng, Greg Okopal,
  Dieter Fox, Byron Boots, and Amir Shaban.
\newblock Terrainnet: Visual modeling of complex terrain for high-speed,
  off-road navigation.
\newblock In \emph{Robotics: Science and Systems}, 2023.

\bibitem[Miki et~al.(2022)Miki, Wellhausen, Grandia, Jenelten, Homberger, and
  Hutter]{miki2022elevation}
Takahiro Miki, Lorenz Wellhausen, Ruben Grandia, Fabian Jenelten, Timon
  Homberger, and Marco Hutter.
\newblock Elevation mapping for locomotion and navigation using gpu.
\newblock In \emph{{IEEE/RSJ} International Conference on Intelligent Robots
  and Systems}, pages 2273--2280. IEEE, 2022.

\bibitem[Moghadam et~al.(2010)Moghadam, Wijesoma, and Moratuwage]{moghadam2010}
Peyman Moghadam, Wijerupage~Sardha Wijesoma, and M.~D.~P. Moratuwage.
\newblock Towards a fully-autonomous vision-based vehicle navigation system in
  outdoor environments.
\newblock In \emph{International Conference on Control Automation Robotics \&
  Vision}, pages 597--602, 2010.
\newblock \doi{10.1109/ICARCV.2010.5707247}.

\bibitem[Nava et~al.(2019)Nava, Guzzi, Chavez-Garcia, Gambardella, and
  Giusti]{nava2019}
Mirko Nava, Jérôme Guzzi, R.~Omar Chavez-Garcia, Luca~M. Gambardella, and
  Alessandro Giusti.
\newblock Learning long-range perception using self-supervision from
  short-range sensors and odometry.
\newblock \emph{IEEE Robotics and Automation Letters}, 4\penalty0 (2):\penalty0
  1279--1286, 2019.
\newblock \doi{10.1109/LRA.2019.2894849}.

\bibitem[Patel et~al.(2024)Patel, Frey, Atha, Spieler, Hutter, and
  Khattak]{patel2024}
Manthan Patel, Jonas Frey, Deegan Atha, Patrick Spieler, Marco Hutter, and
  Shehryar Khattak.
\newblock Roadrunner m\&m -- learning multi-range multi-resolution
  traversability maps for autonomous off-road navigation, 2024.

\bibitem[Qi et~al.(2020)Qi, Mullapudi, Gupta, and Ramanan]{qi2020}
William Qi, Ravi~Teja Mullapudi, Saurabh Gupta, and Deva Ramanan.
\newblock Learning to move with affordance maps.
\newblock In \emph{iclr}, volume abs/2001.02364, 2020.

\bibitem[Ravi et~al.(2024)Ravi, Gabeur, Hu, Hu, Ryali, Ma, Khedr, R{\"a}dle,
  Rolland, Gustafson, Mintun, Pan, Alwala, Carion, Wu, Girshick, Doll{\'a}r,
  and Feichtenhofer]{ravi2024sam2}
Nikhila Ravi, Valentin Gabeur, Yuan-Ting Hu, Ronghang Hu, Chaitanya Ryali,
  Tengyu Ma, Haitham Khedr, Roman R{\"a}dle, Chloe Rolland, Laura Gustafson,
  Eric Mintun, Junting Pan, Kalyan~Vasudev Alwala, Nicolas Carion, Chao-Yuan
  Wu, Ross Girshick, Piotr Doll{\'a}r, and Christoph Feichtenhofer.
\newblock Sam 2: Segment anything in images and videos.
\newblock \emph{arXiv preprint arXiv:2408.00714}, 2024.

\bibitem[Schmid et~al.(2022)Schmid, Atha, Scholler, Dey, Fakoorian, Otsu,
  Ridge, Bjelonic, Wellhausen, Hutter, and Agha-mohammadi]{schmid2022}
Robin Schmid, Deegan Atha, Frederik Scholler, Sharmita Dey, Seyed~Abolfazl
  Fakoorian, Kyohei Otsu, Barry Ridge, Marko Bjelonic, Lorenz Wellhausen, Marco
  Hutter, and Ali-akbar Agha-mohammadi.
\newblock Self-supervised traversability prediction by learning to reconstruct
  safe terrain.
\newblock \emph{{IEEE/RSJ} International Conference on Intelligent Robots and
  Systems}, 2022.

\bibitem[Shaban et~al.(2022)Shaban, Meng, Lee, Boots, and Fox]{shaban2022}
Amirreza Shaban, Xiangyun Meng, JoonHo Lee, Byron Boots, and Dieter Fox.
\newblock Semantic terrain classification for off-road autonomous driving.
\newblock In \emph{Conference on Robot Learning (CORL)}, volume 164, pages
  619--629, 2022.

\bibitem[Shah et~al.(2023)Shah, Sridhar, Dashora, Stachowicz, Black, Hirose,
  and Levine]{shah2023}
Dhruv Shah, Ajay Sridhar, Nitish Dashora, Kyle Stachowicz, Kevin Black, Noriaki
  Hirose, and Sergey Levine.
\newblock Vi{NT}: A foundation model for visual navigation.
\newblock In \emph{Conference on Robot Learning (CORL)}, 2023.

\bibitem[Sridhar et~al.(2023)Sridhar, Shah, Glossop, and Levine]{sridhar2023}
Ajay Sridhar, Dhruv Shah, Catherine Glossop, and Sergey Levine.
\newblock {NoMaD: Goal Masked Diffusion Policies for Navigation and
  Exploration}.
\newblock \emph{OOD Workshop Conference on Robot Learning}, 2023.

\bibitem[Stein et~al.(2018)Stein, Bradley, and Roy]{stein2018}
Gregory~J. Stein, Christopher Bradley, and Nicholas Roy.
\newblock Learning over subgoals for efficient navigation of structured,
  unknown environments.
\newblock \emph{Conference on Robot Learning (CORL)}, 2018.

\bibitem[Wang et~al.(2010)Wang, Zhou, Tu, and Liu]{wang2010}
Mingjun Wang, Jun Zhou, Jun Tu, and Chengliang Liu.
\newblock Learning long-range terrain perception for autonomous mobile robots.
\newblock \emph{International Journal of Advanced Robotic Systems}, 7, 2010.
\newblock \doi{10.5772/7245}.

\bibitem[Wellhausen et~al.(2020)Wellhausen, Ranftl, and Hutter]{wellhausen2020}
Lorenz Wellhausen, Ren{\'e} Ranftl, and Marco Hutter.
\newblock Safe robot navigation via multi-modal anomaly detection.
\newblock \emph{IEEE Robotics and Automation Letters}, 5:\penalty0 1326--1333,
  2020.

\bibitem[Yamauchi(1997)]{yamauchi1997}
Brian Yamauchi.
\newblock A frontier-based approach for autonomous exploration.
\newblock In \emph{Proceedings 1997 IEEE International Symposium on
  Computational Intelligence in Robotics and Automation CIRA'97. 'Towards New
  Computational Principles for Robotics and Automation'}, pages 146--151, 1997.
\newblock \doi{10.1109/CIRA.1997.613851}.

\bibitem[Yang et~al.(2024)Yang, Kang, Huang, Zhao, Xu, Feng, and
  Zhao]{yang2024}
Lihe Yang, Bingyi Kang, Zilong Huang, Zhen Zhao, Xiaogang Xu, Jiashi Feng, and
  Hengshuang Zhao.
\newblock Depth anything v2.
\newblock \emph{arXiv:2406.09414}, 2024.

\bibitem[Zhang et~al.(2023)Zhang, Han, Qiao, Kim, Bae, Lee, and
  Hong]{mobile_sam}
Chaoning Zhang, Dongshen Han, Yu~Qiao, Jung~Uk Kim, Sung-Ho Bae, Seungkyu Lee,
  and Choong~Seon Hong.
\newblock Faster segment anything: Towards lightweight sam for mobile
  applications.
\newblock \emph{arXiv preprint arXiv:2306.14289}, 2023.

\bibitem[Zhang et~al.(2018)Zhang, Chen, Zhang, and He]{zhang2018}
Wei Zhang, Qi~Chen, Weidong Zhang, and Xuanyu He.
\newblock Long-range terrain perception using convolutional neural networks.
\newblock \emph{Neurocomputing}, 2018.

\end{thebibliography}

\clearpage
\appendices
\twocolumn[{
\section{Qualitative Spot Results}
\label{app:spot_gps}

Fig.~\ref{fig:spot_full_gps} shows sample paths each approach took on all courses. As shown there were multiple interventions for the baselines because the robot got off course and needed to be corrected. We also see variations in performance of Traversability + Depth Anything V2 and NoMaD across courses. For example, Traversability + Depth Anything V2 does quite well in the Night course but on Helipad incurs a lot of wandering due to the more open environment.
\vspace{0.5em}

\centering
\includegraphics[width=\textwidth]{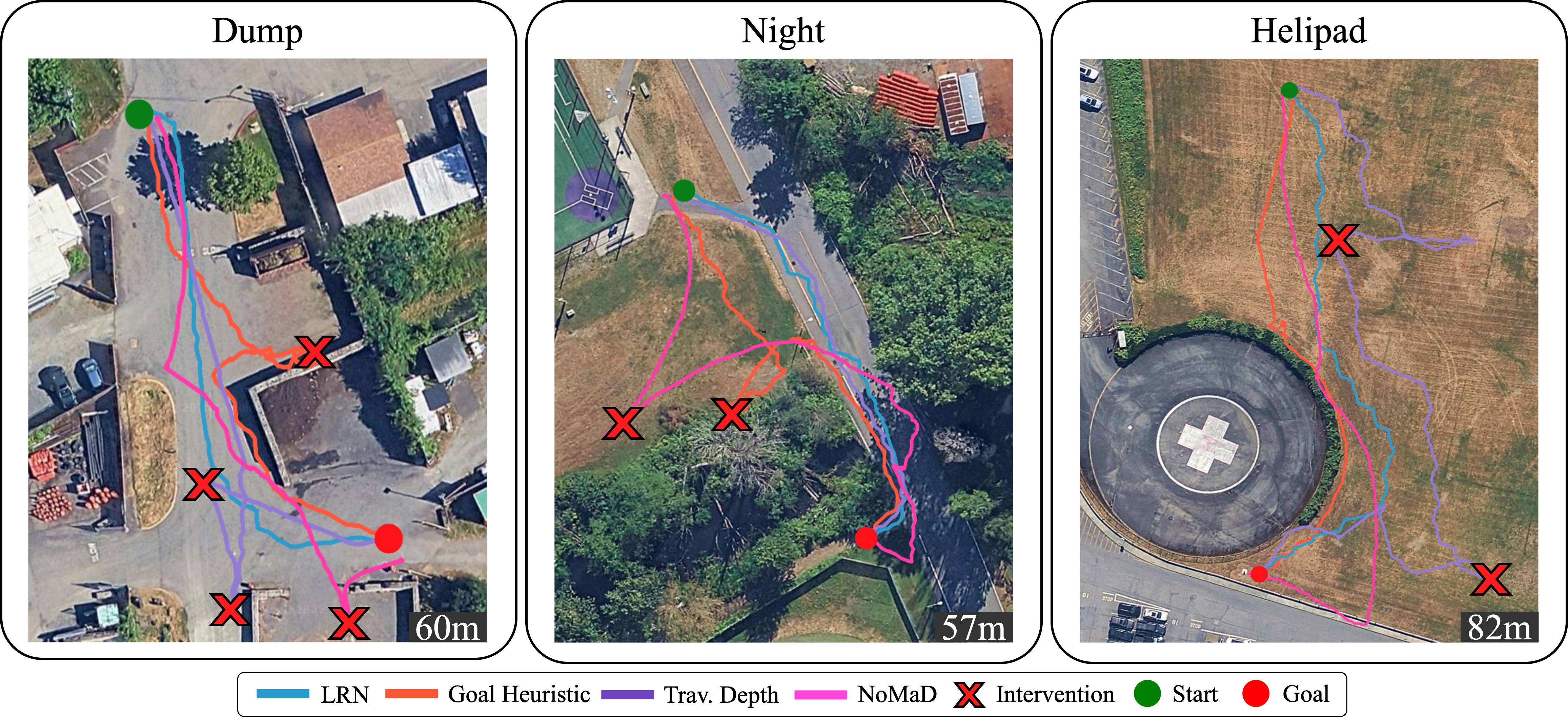}
\captionof{figure}{GPS plots of all approaches on each course. Many of the baselines incurred interventions for going off course and exhibit various degrees of wandering}
\label{fig:spot_full_gps}
\vspace{1em}
}]

\twocolumn[{
\section{Additional Heatmap Predictions}
\label{app:heatmaps}

\subsection{Racer Heavy}

A sample of qualitative heatmap results can be found in Fig.~\ref{fig:racer_qualitative}. Traversability + Depth Anything V2 has varying levels of performance. In the leftmost image, it finds distant hills traversable and thus predicts them as a high score, not considering the uncertainty of getting to the hills. Right of that, it predicts sky as traversable and distant. This happens on and off and is due to fluctuations in depth predictions from Depth Anything V2. In the next two right images, it gets close to the correct hotspots but has no reasoning for whether the robot can continue from that point, thus marking paths leading into dense trees as traversable. 

\texttt{LRN}, on the other hand, gets much closer to the human labels, identifying key openings between trees. While it mostly gets the correct hotspots, it tends to smooth the heat between them more than the true labels, resulting in some heat on undesirable areas.
\vspace{0.5em}

\begin{center}
\includegraphics[width=\textwidth]{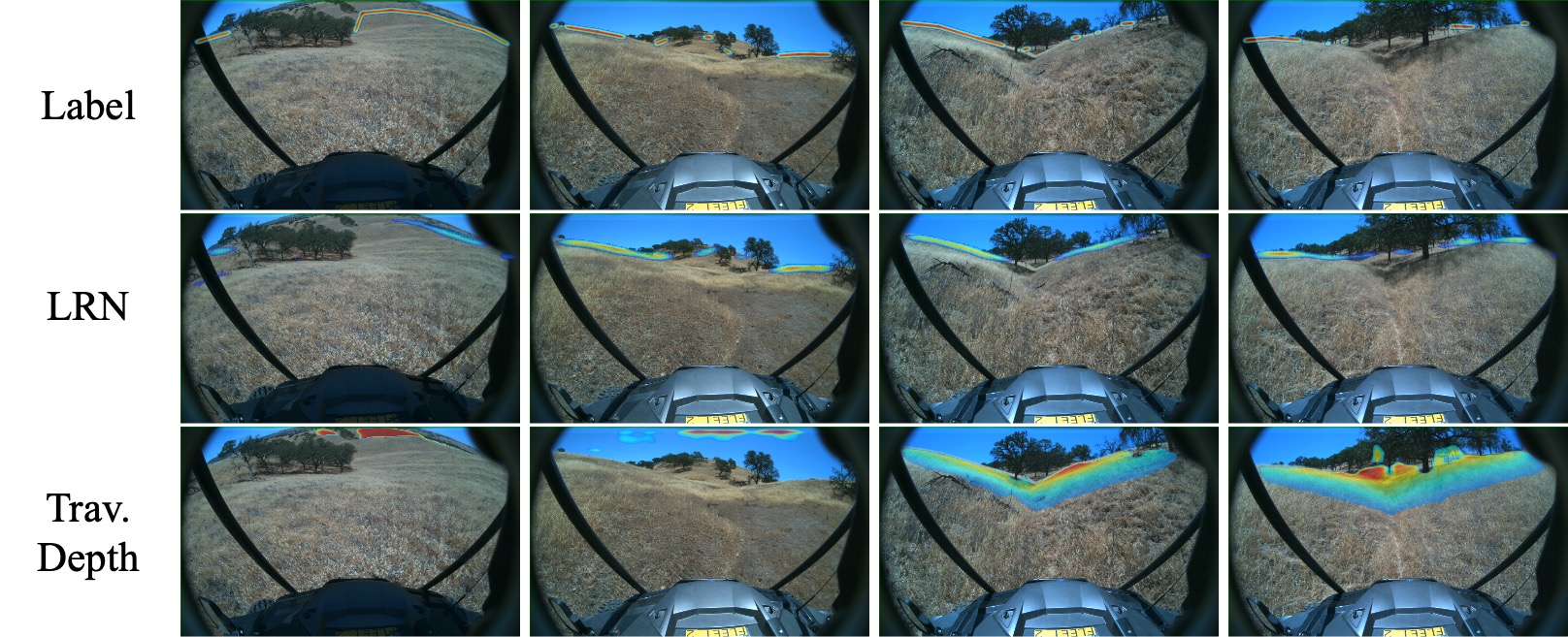}
\captionof{figure}{\textbf{Racer Heavy heatmap predictions} compared to human-labeled heatmaps on the test set.}
\label{fig:racer_qualitative}
\end{center}
}]

\twocolumn[{

\subsection{Spot}

Qualitative heatmap results are presented in Fig.~\ref{fig:spot_offline}. As shown, Traversability + Depth Anything V2 is very sensitive to fluctuations in depth prediction, sometimes giving no hot spots in the heatmap, whereas \texttt{LRN} tended to be more stable. \texttt{LRN} also seemed overly optimistic compared to human labels, which we think contributes to some of the switching behavior in real-world tests.
\begin{center}
\includegraphics[width=\textwidth]{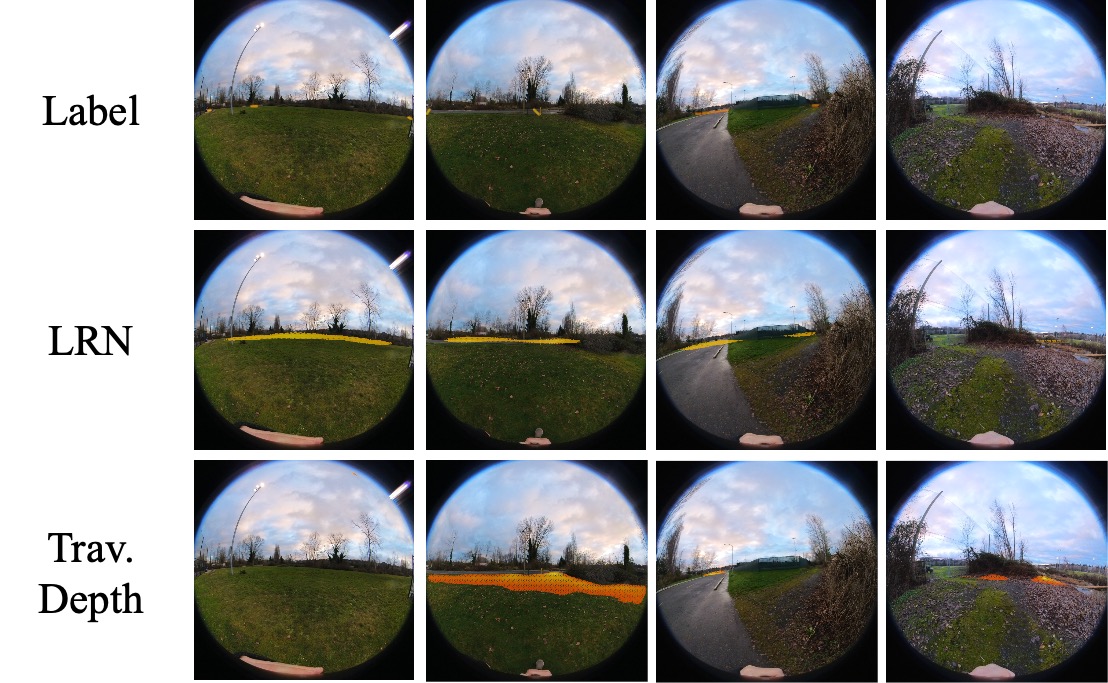}
\captionof{figure}{\textbf{Spot heatmap predictions} compared to human-labeled heatmaps on the test set.}
\label{fig:spot_offline}
\end{center}
}]

\twocolumn[{
\section{Traversability + Depth Anything V2}
\label{app:travdepth}

In this section we explain more about the Traversability + Depth Anything V2 baseline. The heatmaps are created by combining the traversability and monocular depth. We first normalize their individual scores. Depth is only normalized in regions that have a non-zero traversability. We then multiply the scores to produce a heatmap similar to \texttt{LRN} and threshold the values, which are then used instead of \texttt{LRN} hotspots. The monocular depth model we used was Depth Anything V2 base model which was the largest model we could run at a reasonable rate on the Orin AGX.

For Spot traversability, a V-Strong model was not available so we trained a traversability model using the same model and training as \texttt{LRN}, but instead of considering only the hotspot to be 1 in the loss, we mark the whole trajectory as traversable. To improve traversability further we take a trick from V-Strong and expand the traversable region by making a SAM mask seeded from the robot's path.

Fig.~\ref{fig:travdepth} shows the intermediate outputs that lead to the final heatmap scores on Spot. As shown, traversability reasonably covers the space of traversable terrain but emphasizes regions directly in front of the robot, likely due to training trajectories heading straight out. Monocular depth gives reasonable values but becomes foggier further from the robot. Combining and thresholding the two tends to produce heatmaps that resemble \texttt{LRN} (e.g. column 3), but can also be overly optimistic or pessimistic. In practice, the biggest issue was fluctuations in depth prediction, leading to instability in hotspot locations.

\begin{center}
\includegraphics[width=\textwidth]{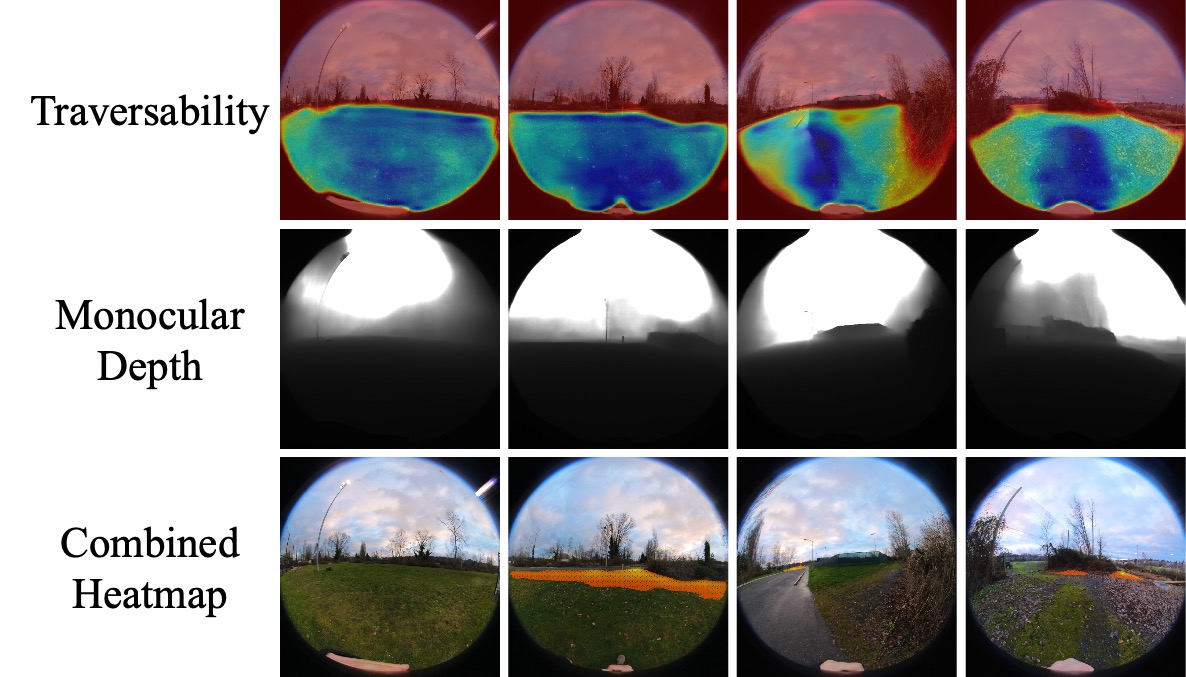}
\captionof{figure}{\textbf{Traversability + Depth Anything V2.} \textcolor{blue}{Blue} indicates more traversable regions, while \textcolor{red}{Red} indicates less traversable areas. Similarly for depth, darker is closer and lighter is further. The resulting heatmap is thresholded to focus on hotspots, but this threshold can be overly conservative, sometimes leading to no hotspots (e.g., column 1).}
\label{fig:travdepth}
\end{center}
}]
 
\end{document}